\documentclass[10pt,twocolumn,letterpaper]{article}

\usepackage[pagenumbers]{cvpr} 

\usepackage{graphicx}
\usepackage{amsmath}
\usepackage{amssymb}
\usepackage{booktabs}
\usepackage{stfloats} 
\usepackage[table,xcdraw]{xcolor} 
\usepackage{multirow} 
\usepackage{setspace} 
\usepackage{colortbl}
\usepackage[table,xcdraw]{xcolor}
\usepackage{bbding} 
\usepackage{xspace} 
\usepackage{colortbl} 
\usepackage{hhline} 

\usepackage[accsupp]{axessibility}  

\usepackage[pagebackref,breaklinks,colorlinks]{hyperref}

\usepackage[capitalize]{cleveref}
\crefname{section}{Sec.}{Secs.}
\Crefname{section}{Section}{Sections}
\Crefname{table}{Table}{Tables}
\crefname{table}{Tab.}{Tabs.}

\definecolor{mr}{rgb}{0.0, 0.0, 0.0}

\def\cvprPaperID{5557} 
\def\confName{CVPR}
\def\confYear{2022}

\newcommand{\age}{AGE\xspace}
\newcommand{\method}{GazeOnce\xspace}
\newcommand{\dataset}{MPSGaze\xspace}

\newcommand*{\affaddr}[1]{#1} 
\newcommand*{\affmark}[1][*]{\textsuperscript{#1}}

\begin{document}


\title{\method: Real-Time Multi-Person Gaze Estimation}
\author{Mingfang Zhang\affmark[1,2],~~Yunfei Liu\affmark[3],~~Feng Lu\affmark[1,3,*]\\
\affaddr{\affmark[1]Peng Cheng Laboratory,~~\affmark[2]The University of Tokyo,\\\affmark[3]State Key Lab. of VR Technology and Systems, School of CSE, Beihang University}\\
{\tt\small mfzhang@iis.u-tokyo.ac.jp,~~lyunfei@buaa.edu.cn,~~lufeng@buaa.edu.cn}
}

\maketitle

\begin{abstract}
\vspace{-0.7em}Appearance-based gaze estimation aims to predict the 3D eye gaze direction from a single image. While recent deep learning-based approaches have demonstrated excellent performance, they usually assume one calibrated face in each input image and cannot output multi-person gaze in real time. However, simultaneous gaze estimation for multiple people in the wild is necessary for real-world applications. In this paper, we propose the first one-stage end-to-end gaze estimation method, \method, which is capable of simultaneously predicting gaze directions for multiple faces ($>$10) in an image. In addition, we design a sophisticated data generation pipeline and propose a new dataset, \dataset, which contains full images of multiple people with 3D gaze ground truth. Experimental results demonstrate that our unified framework not only offers a faster speed, but also provides a lower gaze estimation error compared with state-of-the-art methods. This technique can be useful in real-time applications with multiple users.

   \let\thefootnote\relax\footnotetext{*\hspace{0.1em}Corresponding author.}
   \let\thefootnote\relax\footnotetext{This work is done during M. Zhang's internship at Pengcheng Lab.}
   \let\thefootnote\relax\footnotetext{Accepted to CVPR 2022.}
\end{abstract}


\section{Introduction}
\label{sec:intro}

Eye gaze is one of the important channels in revealing human intentions. It has been adopted for a wide range of applications such as human-computer interaction \cite{wang2015hybrid}, virtual/augmented reality \cite{burova2020utilizing, wang2020comparing}, medical diagnostics \cite{castner2020deep}, and surveillance systems \cite{marois2021improving}. To estimate the gaze direction, various systems have been developed. However, fast and accurate calculation of gaze direction in a large range of environment remains challenging.

\begin{figure}[t]
  \centering
  \includegraphics[width=\linewidth]{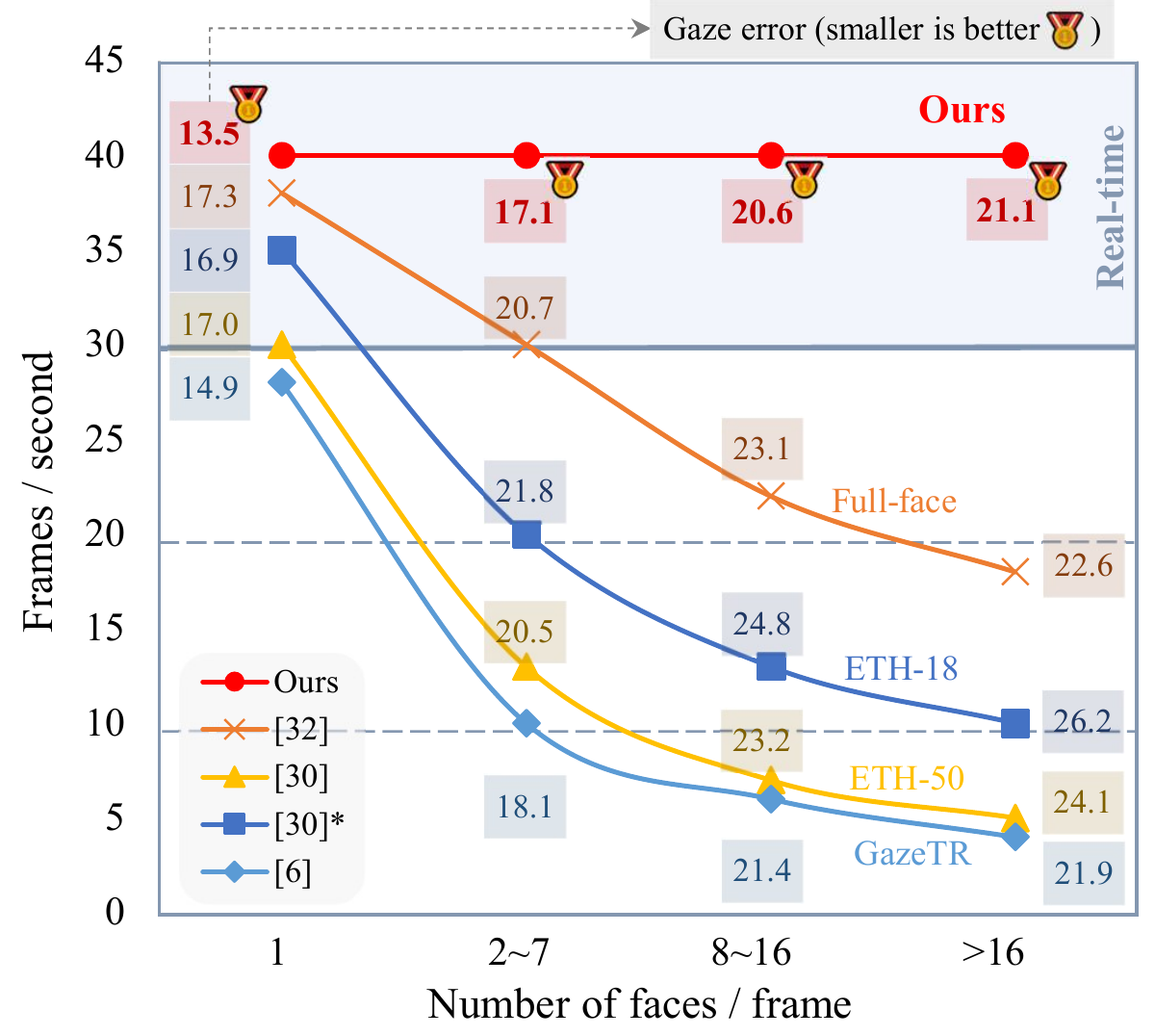}
  \vspace{-2em}
  \caption{Compared with previous appearance-based gaze estimation methods \cite{zhang2020eth,zhang2017s,cheng2021gaze}, our method is the only one that can maintain the real-time speed as the number of faces increases in the input image. Consider the average gaze accuracy across different face resolutions, our method also achieves the best performance. The experiment setting is the same as \cref{tab:tradition}.}
  \label{fig:first1}
  \vspace{-0.8em}
\end{figure}

\begin{figure*}[t]
  \centering
  \includegraphics[width=\linewidth]{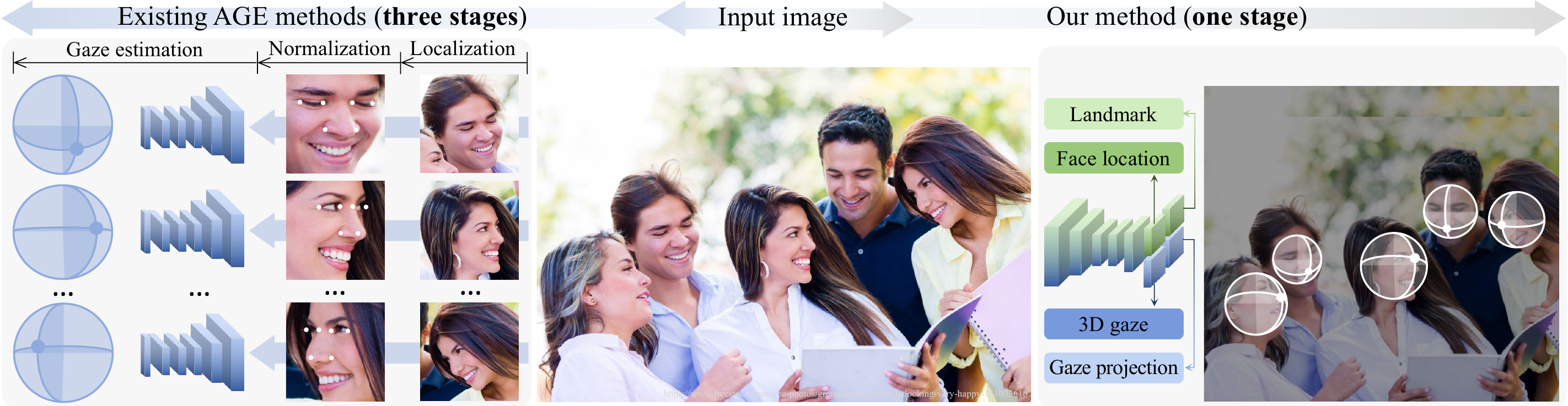}
  \caption{Comparison between existing appearance-based gaze estimation (\age) methods and our method. \age methods usually conduct localization, normalization, and gaze estimation for each face one by one. We present the first one-stage method to simultaneously estimate gaze directions for multiple people in one pass.}
  \label{fig:first2}
  \vspace{-1em}
\end{figure*}

With the development of deep learning, appearance-based gaze estimation has attracted more and more attention, \ie, gaze estimation using face images captured by common cameras. The main drawbacks of existing methods are: 1) they usually only support the gaze estimation for a single person, while multi-person with different head poses have been  less explored; 2) they need to pre-process the images, \ie, cropping and calibration of the face images, resulting in a longer computation time. Fig. \ref{fig:first2} illustrates the typical flow of existing systems. It first extracts face ROI using a face detector, calibrates each face using the detected facial landmarks, and then the normalized face is fed into the gaze direction estimation system. It can be seen that the system errors accumulated after these steps. Moreover, their computational complexity is proportional to the number of people in the image, and they normally cannot operate in real time when there are more than 5 faces in each frame, as shown in \cref{fig:first1}.

In this paper, we reframe the multi-person gaze estimation as a single-stage regression task, which directly maps image pixels to multiple gaze directions. Specifically, we propose the first one-stage gaze estimation method, \ie, \method, which estimates all human gaze directions within one pass. The proposed method not only estimates gaze directions but also predicts auxiliary face information including bounding box and facial landmarks. In addition, we carefully design a projection-based self-supervision loss for 3D gaze estimation.

Another difficulty to overcome is about the dataset. Appearance-based gaze estimation relies on high-quality datasets with face images and ground truth gaze directions. However, obtaining gaze ground truth is very challenging. Many gaze datasets have been released~\cite{zhang2020eth,zhang2017s}, while they usually record data for each single person in a strictly controlled environment, leading to limited image styles and body poses. On the other hand, manual annotation of 3D gaze directions is time-consuming and error-prone.

In order to train our \method method, a new high quality dataset is needed with multiple people and their gaze ground truth in every  image. To this end, we propose a sophisticated gaze swapping method for generating a high-quality multi-person gaze dataset. The proposed \dataset dataset has no restrictions on the number of people and scenes, and is also easily extensible. This makes the training and evaluation of multi-person gaze estimation possible.

Based on the proposed dataset, our method not only achieves real-time multi-person gaze estimation, but also outperforms state-of-the-art methods in terms of estimation error and running time, as shown in \cref{fig:first1}.

In summary, our main contributions are as follows:
\begin{itemize}
\vspace{-0.5em}
\item {We propose the first one-stage gaze estimation method, \ie, \method, which can estimate multi-user gaze directions simultaneously in a single image. In addition, we design a projection-based self-supervised strategy that can further improve the gaze accuracy.}
\vspace{-0.5em}
\item {We provide a new gaze dataset \dataset, which enables one-stage gaze estimation training and evaluation. This dataset is generated by a sophisticated swap-gaze procedure to produce full images of multi-person with their gaze ground truth.}
\item {Our method outperforms state-of-the-art methods in terms of gaze accuracy and speed, especially in the cases of a large number of faces.}

\end{itemize}

\section{Related Works}
\label{sec:relatedworks}

\vspace{-0.5em}

\paragraph{Appearance-based gaze estimation (\age).} AGE has been a long-standing computer vision problem \cite{lu2014adaptive,lu2014learning}. Recent deep learning-based \age methods \cite{zhang2015appearance,zhang2017s,cheng2020gaze,fischer2018rt} have significantly improved the accuracy using various strategies, such as a coarse-to-fine approach \cite{cheng2020coarse}, an adversarial learning approach \cite{wang2019generalizing}, a self-attention approach \cite{bao2021adaptive}, \etc. At the same time, large-scale gaze datasets \cite{zhang2020eth,kellnhofer2019gaze360,fischer2018rt,sugano2014learning} have been proposed. Most of them are collected in laboratory environments with a strict setting of multi-view cameras, 3D positions of human participants and gaze targets, \etc.
This procedure always results in these datasets containing only single face images in a limited number of scenes.
Correspondingly, current \age methods all assume that there is only one calibrated face in the input image. However, this will lead to a disadvantage that the speed of current \age methods depends on the number of faces in the input image. Most methods cannot achieve real-time performance when there are multiple people in the image.

\vspace{-1em}

\paragraph{Real-time multi-face process.} Face understanding receives keen attention because of its wide range of applications. Many methods have been proposed for face localization \cite{najibi2017ssh}, facial expression recognition \cite{yang2018facial}, head pose estimation \cite{albiero2021img2pose}, \etc. With the development of object detection methods \cite{liu2016ssd}, one-stage methods for multi-face understanding are favored by real-time applications because of their lightweight design and high accuracy. For example, face detection methods \cite{najibi2017ssh,deng2020retinaface} apply the one-stage architecture and design more efficient modules for face characteristics. Correspondingly, large-scale face datasets \cite{yang2016wider} have been constructed by employing a large number of manual annotations. In addition, researchers find that it is an efficient method \cite{zhang2016joint,ranjan2017hyperface} to conduct multi-task (landmark, head pose, gender, \etc) learning alongside face detection because these tasks share common facial features. Inspired by these works, we propose to develop a one-stage gaze estimation method.

\vspace{-0.3em}
\section{Multi-Person Swap Gaze Dataset}
\label{sec:swapfacedataset}
\vspace{-0.3em}

\begin{figure*}[t]
  \centering

  \includegraphics[width=\linewidth]{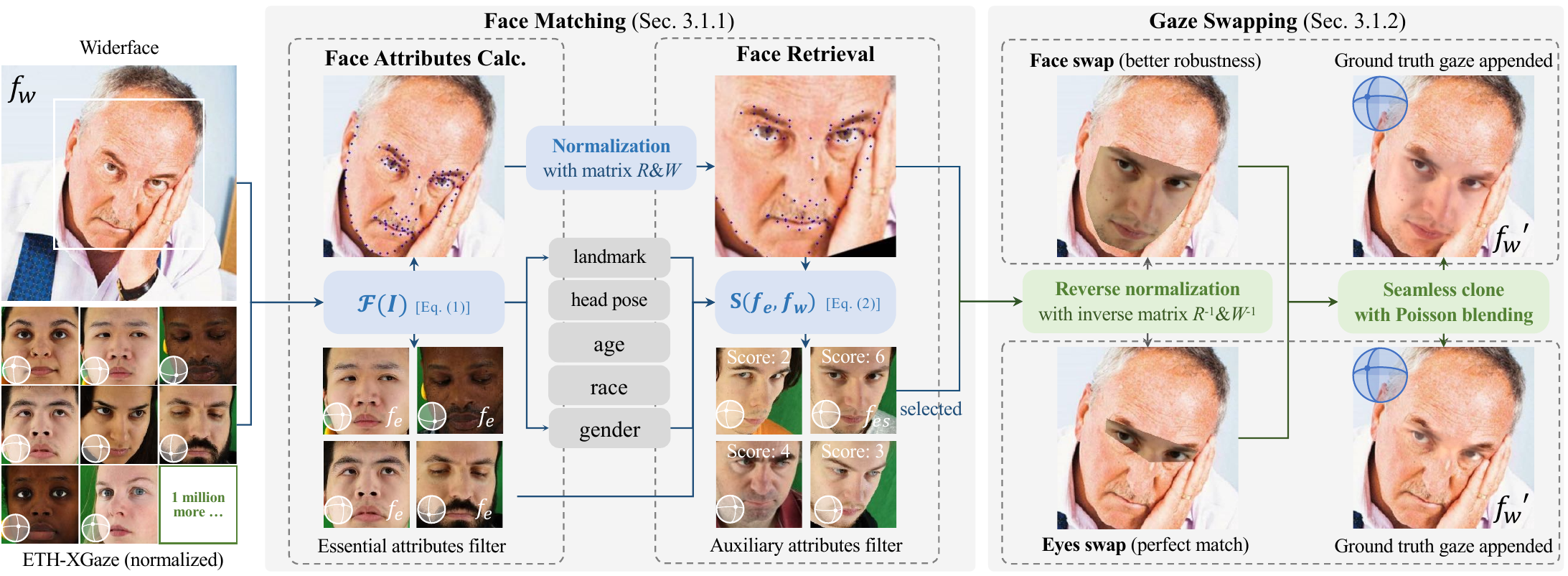}
  \caption{Generation of the \dataset. To create a dataset with full images of multi-person with gaze ground truth, we conduct gaze swapping between the Widerface \cite{yang2016wider} dataset (with face bounding box labels) and the ETH-XGaze\cite{zhang2020eth} dataset (with gaze labels). The pipeline consists of 2 phases, matching and swapping. For each qualified face in Widerface, we retrieve the nearest face from ETH-XGaze by filtering various face attributes. Based on the matching result, we design 2 strategies to swap gaze, \ie, face exchange and eyes exchange.}
  \label{fig:matchpipe}
\end{figure*}

We propose a new Multi-Person Swap Gaze Dataset, \dataset, for our task of gaze estimation for multiple people in one stage. To the best of our knowledge, existing datasets either only contains face information (\eg, bounding box, landmark, \etc) or contains normalized single faces with gaze labels. Therefore, our first obstacle is to construct a dataset that contains both multiple people in the wild and corresponding gaze ground truth. To this end, we propose to merge the advantages of face datasets and gaze datasets to enable the training and evaluation of multi-person gaze estimation in one stage. 
In the following, we first introduce the pipeline of generating the \dataset dataset, then show the details of the dataset.

\subsection{Generation Pipeline}
We choose the largest and most common gaze dataset available, ETH-XGaze \cite{zhang2020eth}, and the face detection dataset, Widerface \cite{yang2016wider}, for our task. The proposed approach consists of two phases, face matching and gaze swapping, as shown in \cref{fig:matchpipe}.

\vspace{-1em}
\subsubsection{Face Matching}

The left part of \cref{fig:matchpipe} shows the process of face matching between the two datasets. First, we conduct face attributes calculation for each qualified face in Widerface \cite{yang2016wider} and ETH-XGaze \cite{zhang2020eth} by
\begin{equation}
    \mathbf{A} = \mathcal{F} (I), 
\end{equation}
where $\mathbf{A}$ is the attributes extracted from a single-face image $I$. Here $\mathbf{A} = \{\textbf{a}_{lmk}, \textbf{a}_{pose}, \textbf{a}_{age}, \textbf{a}_{race}, \textbf{a}_{gender}\}$ and $\textbf{a}_{lmk}\in \mathbb{R}^{68\times 2}$, $\textbf{a}_{pose}\in \mathbb{R}^{2}$, $\textbf{a}_{age}\in \mathbb{R}^{9}$, $\textbf{a}_{race}\in \mathbb{R}^{7}$, $\textbf{a}_{gender}\in \mathbb{R}^{2}$. The function $\mathcal{F}$ is implemented by state-of-the-art methods~\cite{Wang_2019_ICCV,karkkainenfairface}.

Next, for each qualified face $f_w$ in the Widerface \cite{yang2016wider}, we retrieve the nearest face $f_{es}$ from the ETH-XGaze \cite{zhang2020eth}.
The implementation of our retrieval is as follows. 
1) We first conduct an essential-attribute (we choose gender) filter for faces in ETH-XGaze \cite{zhang2020eth} to match with $f_w$. The chosen faces are called $f_{e}$. 
2)
$f_w$ is normalized according to its landmarks and the normalization steps are consistent with ETH-XGaze \cite{zhang2020eth}.
3) We save the image warp matrix \textit{W} and the head pose rotation matrix \textit{R}. 
4) We calculate the difference in landmarks and head poses between $f_w$ and faces in $f_{e}$. Jointly we compute a matching score for each face in $f_{e}$ by a scoring function
\begin{equation} \label{eq:matching_loss}
\mathcal{S}(f_e,f_w) = \sum_{\tau \in \{lmk,pose\}} \alpha_{\tau} * |\textbf{a}_{\tau,w} - \textbf{a}_{\tau,e}|,
\end{equation}
where $\alpha_{\tau}$ is determined by the experience of comparing the matching results.
5) We keep $n$ highest-scoring faces for the final filtering, auxiliary-attribute filtering, where we penalize the score of the left $n$ faces by a joint measurement of age and race differences. 
6) Finally, we choose the face $f_{es}$ with the final highest score in $f_{e}$ as a match for $f_w$ from Widerface \cite{yang2016wider}.

\begin{figure}[t]
  \centering
  \includegraphics[width=\linewidth]{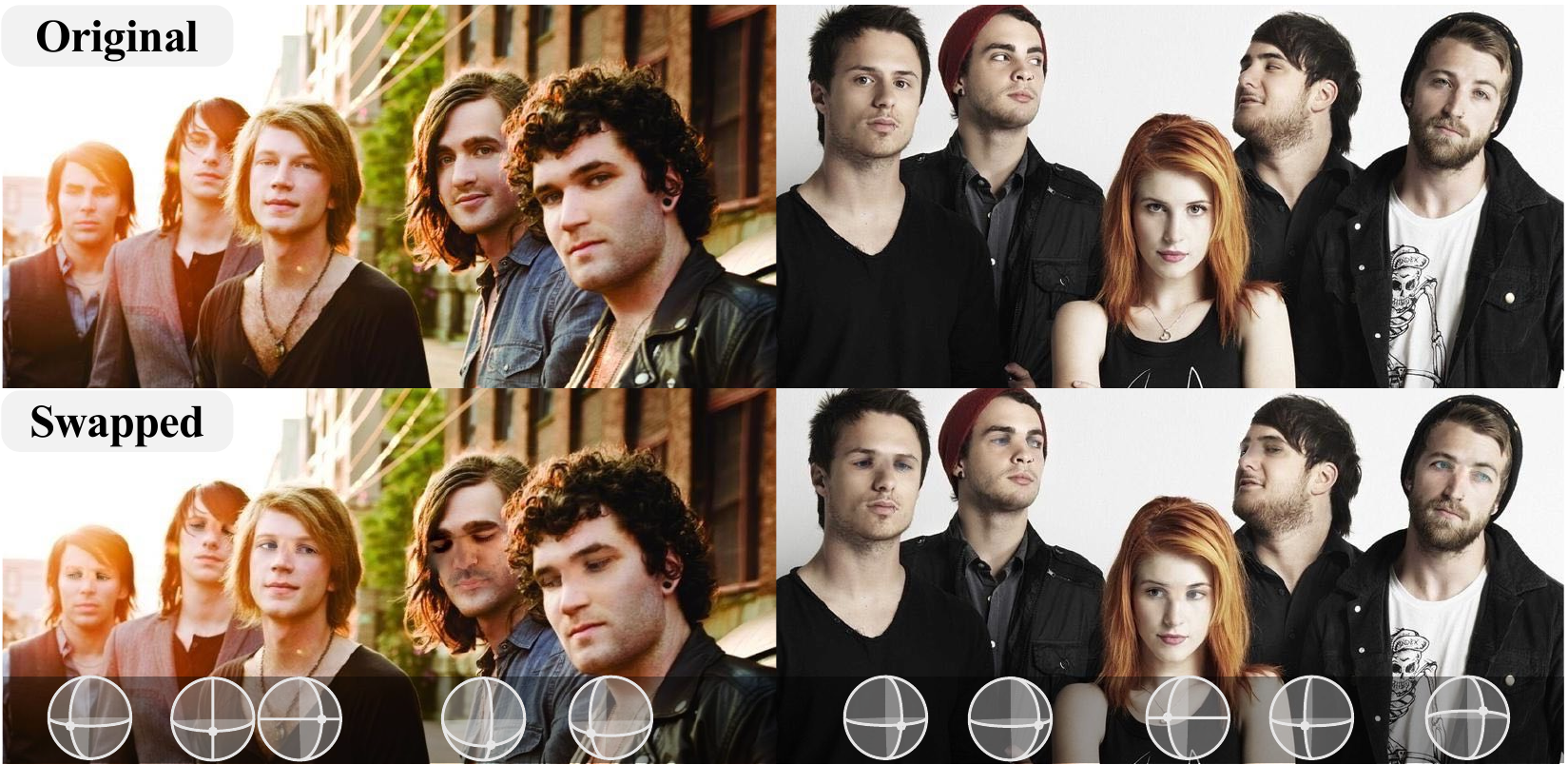}
  \caption{ Gaze swap results (second row) with ground truth gaze labels in our \dataset dataset. Please zoom in for details.}
  \label{fig:swap-im}
  \vspace{-1em}
\end{figure}

\vspace{-1em}

\subsubsection{Ground Truth Available Gaze Swapping}
\vspace{-0.5em}

We propose a gaze swapping method to produce synthetic face images $f_w'$ with ground truth gaze $g'$. Through affine transformations,
the ground truth gaze of $f_{es}$ can be preserved. The procedure is shown in the right part of \cref{fig:matchpipe}.

Given a pair of matched faces $f_w$ and $f_{es}$, we swap the gaze as follows.
1) We first warp $f_{es}$ to match with the original $f_w$ through \textit{W}$^{-1}$ and \textit{R}$^{-1}$, where \textit{W} and \textit{R} are image warp matrix and head pose rotation matrix. These two matrices are pre-computed in the step of Face Matching. 
2) Then we load the matching error computed by \cref{eq:matching_loss} according to the landmarks and head poses difference between the face pair. 
3) When the error is under a given threshold, we only swap the eye region by replacing $f_w$ with $f_{es}$. 
4) Otherwise, we keep the whole face region of $f_{es}$ to replace $f_w$. This operation produces more robust results because the head pose information of face $f_{es}$ is preserved. 
5) Next, we employ the Poisson blending method~\cite{perez2003poisson} to seamlessly fuse the two overlapped faces for the sake of reality.
6) Finally, we apply the inverse rotation matrix \textit{R}$^{-1}$ on the original gaze label of $f_{es}$ as the ground truth gaze $g'$ for the final swapped image~$f_w'$.

\subsection{Details of \dataset}

In the Widerface \cite{yang2016wider} training dataset, 24282 faces are swapped leaving faces that are too small, too blurry, and with too much occlusion unswapped. Some examples are shown in \cref{fig:swap-im}. As shown in \cref{tab:datasets}, compared to other gaze datasets, the \dataset contains images of $\sim$20 thousand people subjects with multiple faces per image. In addition, \dataset shares the advantages of \cite{yang2016wider} that it contains images in a large variety of wild scenes which are specially designed.

\section{Method}
\label{sec:method}

\begin{figure*}[t]

  \centering
  \includegraphics[width=\linewidth]{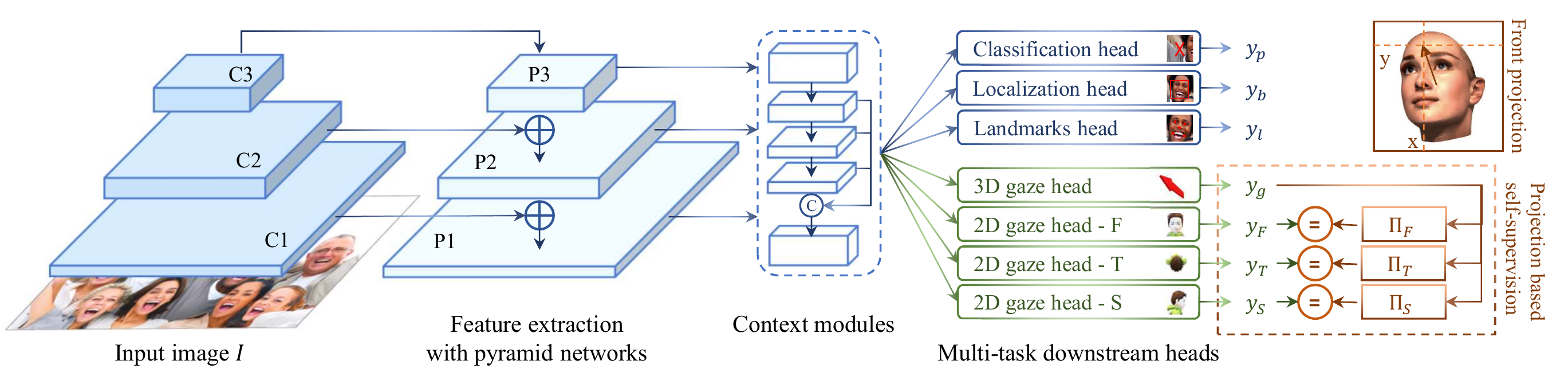}
  \caption{Overview of the \method framework. The feature extraction is based on feature pyramids followed by context modules \cite{najibi2017ssh}, which is adopted from RetinaFace \cite{deng2020retinaface}. Next, we calculate a joint loss for gaze estimation and face localization for each positive anchor. To achieve higher gaze accuracy, we propose to project the 3D gaze from 3 directions as an auxiliary supervision signal and design a self-supervision loss function to constrain the predictions from different views to be equal.}
  \label{fig:pipe}
  \vspace{-1em}
\end{figure*}

After acquiring the \dataset dataset which enables the training of one-stage multi-person gaze estimation, we propose the \method framework. The architecture overview is illustrated in \cref{fig:pipe}. The input to our model is a full image containing any number of faces and the output is multi-user gaze directions. Instead of processing every face one by one, we propose the first model to estimate gaze for multiple people in one stage.

\subsection{Multi-task Learning for Face Detection and Gaze Estimation}

We equip the proposed \method with a multi-task learning strategy, \ie, jointly optimizing face localization and gaze estimation. 
Inspired by the RetinaFace \cite{deng2020retinaface}, we employ a similar architecture for our task. The proposed \method mainly consists of two components: \textit{feature extractor} and \textit{downstream heads}. 
The \textit{feature extractor} aims to encode different faces from the input image $I$ into latent codes. To get a rich embedding in which faces with different scales can be treated equally, we adopt the feature pyramids and the context modules from~\cite{deng2020retinaface} as the feature extractor. Specifically, different levels of the feature pyramid produce features of different scales computed from the output of the corresponding stages of MobileNet \cite{sandler2018mobilenetv2} or ResNet \cite{he2016deep} using top-down and lateral connections.
Next, for each feature level, context modules \cite{najibi2017ssh} are implemented to increase the receptive field. 
The feature extraction is effective for both the face detection task and the gaze estimation task because they share all-face information, such as the head pose, besides the eye-region information.
Similar conclusions have also been made in \cite{zhang2017s}.

\begin{table}[t]
\centering
\small
\begin{tabular}{l|ccc}
\toprule
Gaze dataset & \# people & \# faces/image & Unconstrained\\ 
\midrule
MPIIGaze \cite{zhang2017s}  & 15  & 1    & $\times$    \\
ETH-XGaze \cite{zhang2020eth} & 110 & 1    & $\times$    \\
Gaze360 \cite{kellnhofer2019gaze360}   & 238 & 1    & $\times$ \\
\dataset~(ours)  & $\sim$10$^4$  & 1$\sim$30 &  \checkmark   \\
\bottomrule
\end{tabular}%
\caption{Comparison with other gaze datasets. In existing datasets, limited number of \textit{people} subjects are requested to look at preset targets to collect \textit{constrained} gaze data.}
\label{tab:datasets}
\end{table}

After the feature extraction, 1$\times$1 convolutions are used as \textit{downstream heads} for different tasks. 
For the face detection task, we employ three heads, namely classification head, localization head, and landmarks head. These three heads are used to predict probabilities of existence $y_p$, bounding boxes $y_b$, and positions of landmarks $y_l$, respectively. 
We design a 3D gaze head and three auxiliary projection heads for the gaze estimation task. These heads estimate the 3D gaze $y_g$ and three 2D gaze projections $y_F$, $y_T$, and $y_S$.
For each training anchor $i$, we minimise a multi-task loss:
\begin{equation} \label{loss:multi_task}
    \mathcal{L} = \alpha \mathcal{L}_{face} + \beta \mathcal{L}_{gaze},
\end{equation}
where:
\begin{equation}
\begin{aligned}
\mathcal{L}_{face} = &\mathcal{L}_{class}(y_p^i,y_p^{i*}) + \lambda_1 y_p^{i*}\mathcal{L}_{box}(y_b^i,y_b^{i*})
\\ &+ \lambda_2 y_p^{i*}\mathcal{L}_{landmark}(y_l^i,y_l^{i*}),
\end{aligned}
\label{eq:lface}
\end{equation}
where notations with $*$ are the corresponding ground truth. Landmark regression is an auxiliary task to benefit face detection which is proven in \cite{zhang2016joint,deng2020retinaface}. We will introduce $\mathcal{L}_{gaze}$ in the next section.

\begin{figure}[b]
\vspace{-1em}
  \centering
  \includegraphics[width=0.75\linewidth]{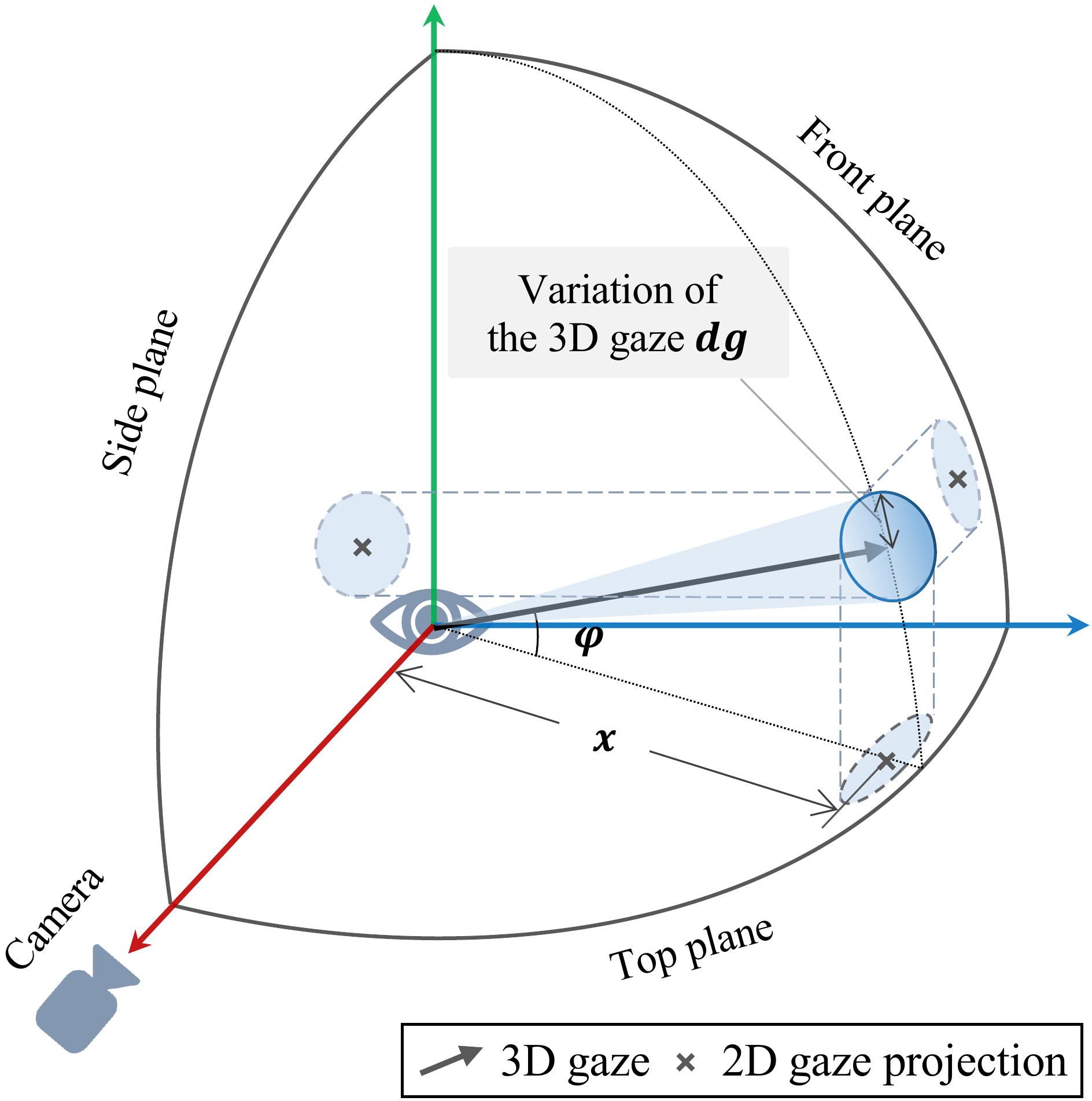}
  \caption{Visualization of 2D projections of a 3D gaze onto 3 planes. When there is a variation $d g$ in the 3D gaze, its projection variations in pixels are different on the three planes.}
  \label{fig:proj}
\end{figure}

\subsection{2D Projection-based Self-supervision for 3D Gaze Estimation}
\label{sec:proj}

We propose a projection-based self-supervision technique for our method. The idea is to project the 3D gaze onto three planes to form three 2D gaze estimation sub-tasks, which constrains the original 3D gaze estimation task in a self-supervised manner.

Inspired by 3D head pose estimation work \cite{albiero2021img2pose} which projects the 3D pose onto the image plane for supervision, we apply the projection operation in the 3D gaze estimation task. However, we notice that projecting a 3D gaze onto different planes may result in different performances.

\cref{fig:proj} shows three projections of a 3D gaze onto three planes, namely the front plane (image plane), the side plane, and the top plane. 
When there is a certain variation $dg$ in the 3D gaze, its projection variations in pixels are different on the three planes.
For instance, the 2D projection point on the side plane falls near the origin, and thus the corresponding pixel variation is larger.

To mathematically model such differences, we introduce the concept of \textit{2D Gaze Sensitivity (GS)}:
\begin{equation}
\begin{aligned}
&GS = \frac{dg}{dx} =\frac{d\varphi}{dx} = \frac{r}{\sqrt{r^2-x^2}}.
\end{aligned}
\label{eq:gr}
\end{equation}
GS defines the consequential variation $dg$ of the 3D gaze angle with respect to a change $dx$ at the position $x$ on the projection plane. According to \cref{eq:gr}, the further $x$ is apart from the origin, the larger the 2D GS is.

Clearly, lower 2D GS means lower uncertainty in the corresponding 3D gaze direction given the fixed pixel resolution, which is good for our task. However, as shown in \cref{fig:proj}, the 2D GS can be large in two projection planes with large $x$ values. Therefore, we propose to use all of the three projections on different planes to ensure the existence of at least one lower 2D GS.

We implement this idea in our network. As shown in \cref{fig:pipe}, besides the 3D gaze task, we introduce three additional sub-tasks to estimate three sets of 2D gaze points on the projection planes, namely $y_F$, $y_T$ and $y_S$, where `F', `T' and `S' stand for front, top, and side. 

Then, a self-supervised mechanism can be constructed by checking the consistency between each of the three 2D gaze outputs and three projections $\Pi_{F}$, $\Pi_{T}$ and $\Pi_{S}$ of the 3D gaze output $y_g$ respectively as shown in \cref{fig:pipe}. Specifically, the three projections of $y_g$ follow these equations:
\begin{equation}
\begin{aligned}
&\Pi_{F}(\theta,\phi) = [-r * \sin{\phi} * cos{\theta}, &-r * \sin{\theta}],\\
&\Pi_{T}(\theta,\phi) = [r * \cos{\phi} * cos{\theta} , &-r * \sin{\phi} * cos{\theta}],\\
&\Pi_{S}(\theta,\phi) = [-r * \cos{\phi} * cos{\theta}, &-r * \sin{\theta}], 
\end{aligned}
\label{eq:fproj}
\end{equation}
where $\Pi$ is the projection function, $r$ is the half face width, and $\{\theta$, $\phi\}$ are the \textit{pitch} and \textit{yaw} components of $y_g$. 

As explained above, by projecting the 3D gaze onto three planes simultaneously, there is always at least one projection with low 2D GS, which is favorable for the estimation.
This is supported by the results in \cref{tab:ablation}.

\paragraph{Loss Design.} By constraining the three 2D gaze estimation points $y_F,y_T,y_S$ on the three planes to be equal to the three projections of the 3D gaze estimation direction $y_g$, our self-supervision improves the final gaze estimation accuracy. The self-supervision is implemented by using the following loss function:
\begin{equation}
\begin{aligned}
\mathcal{L}_{self} = \sum_{\tau \in \{F,T,S\}} |y_\tau - \Pi_\tau(y_g)|_1*e^{-p_\tau}+p_\tau,
\end{aligned}
\label{eq:lself}
\end{equation}
where $F, T, S$ represents front, top, and side, $\Pi$ functions are from \cref{eq:fproj}, and $p$ is trainable weights \cite{garau2021deca} associated with each projection plane. Finally, the total loss is $\mathcal{L} = \alpha \mathcal{L}_{face} + \beta \mathcal{L}_{gaze}$, where $\mathcal{L}_{face}$ is defined in \cref{eq:lface} and $\mathcal{L}_{gaze}$ is defined as:
\begin{equation}
\begin{aligned}
\mathcal{L}_{gaze} &= \lambda_1 \mathcal{L}_{self} + \lambda_2 \mathcal|y_g - y_g^*|_1\\
&+ \lambda_3 \sum_{\tau \in \{F,T,S\}} |y_\tau - \Pi_\tau(y_g^*)|_1,
\end{aligned}
\end{equation}
where $*$ represents the ground truth and other expressions are the same as in \cref{eq:lself}.


\section{Experiments}

\begin{table*}
\centering
\caption{Running speed comparison on NVIDIA GeForce RTX 2080. Our one-stage gaze estimation method can run at almost the same speed as RetinaFace \cite{deng2020retinaface} (SOTA face detection method). Assuming that existing AGE methods \cite{zhang2017s,zhang2020eth,cheng2021gaze} employ RetinaFace \cite{deng2020retinaface} for face detection costing time $T$ (25ms on average), their average values of running speed tested on the Widerface \cite{yang2016wider} validation set are shown in the table.}
\begin{spacing}{1.15}
\arrayrulecolor[rgb]{0.902,0.902,0.902}
\begin{tabular}{|cccccc|} 
\hline
Method & Ours (MobileNet) & Full-face \cite{zhang2017s} & ETH-18 \cite{zhang2020eth} & ETH-50 \cite{zhang2020eth} & GazeTR \cite{cheng2021gaze} \\
\rowcolor[rgb]{0.902,0.902,0.902} Time/image (ms) & 24.93 & $T (\approx 25)$+1.21$\times$\textit{\#face} & $T$+3.15$\times$\textit{\#face} & $T$+6.64$\times$\textit{\#face} & $T$+9.98$\times$\textit{\#face} \\
\hline
\end{tabular}
\end{spacing}
\label{tab:speed}
\end{table*}

\begin{table*}[t]
\centering
\caption{Gaze error evaluated on the \dataset. It shows comparison between our method and existing AGE methods, including Full-face \cite{zhang2017s}, ETH-18 \cite{zhang2020eth}, ETH-50 \cite{zhang2020eth}, and GazeTR \cite{cheng2021gaze} (trained on ETH-XGaze dataset \cite{zhang2020eth}). Our method shows higher accuracy in grading comparisons of faces with various scales, even when compared to the transformer-based method \cite{cheng2021gaze}.}
\begin{spacing}{1.15}
\arrayrulecolor[rgb]{0.902,0.902,0.902}
\begin{tabular}{|ccc||cccccccc|} 
\hline
\multirow{2}{*}{Method} & \multirow{2}{*}{Backbone} & \multirow{2}{*}{Input} & \multicolumn{8}{c|}{Gaze error (lower is better) \wrt \textbf{the width of faces}} \\ 
\cline{4-11}
 &  &  & \small{30-60} & \small{60-90} & \small{90-120} & \small{120-150} & \small{150-180} & \small{180-210} & \small{210-240} & \small{\textgreater240} \\ 
\hline
\rowcolor[rgb]{0.902,0.902,0.902} Full-face & AlexNet & \footnotesize{1 normalized face} & 24.99 & 20.00 & 17.56 & 17.03 & 16.47 & 14.74 & 13.43 & 12.31 \\
ETH-18 & ResNet18 & \footnotesize{1 normalized face} & 28.89 & 21.93 & 16.66 & 14.90 & 14.33 & 12.44 & 11.68 & 10.32 \\
\rowcolor[rgb]{0.902,0.902,0.902} ETH-50 & ResNet50 & \footnotesize{1 normalized face} & 29.82 & 21.87 & 16.93 & 14.76 & 13.87 & 11.79 & 11.13 & 9.98 \\
GazeTR & ResNet18 & \footnotesize{1 normalized face} & 24.51 & \underline{16.84} & 14.59 & 13.37 & 13.65 & 11.72 & 10.71 & 9.96 \\
\rowcolor[rgb]{0.902,0.902,0.902} Ours & \footnotesize{MobileNet0.25}  & 1 full image & \underline{22.94} & 17.55 & \underline{13.69} & \underline{11.08} & \underline{11.13} & \underline{9.41} & \underline{8.17} & \underline{7.74} \\
Ours & ResNet50 & 1 full image & \textbf{21.17} & \textbf{13.77} & \textbf{10.58} & \textbf{7.9} & \textbf{8.57} & \textbf{6.68} & \textbf{6.01} & \textbf{5.56} \\
\hline
\end{tabular}
\arrayrulecolor{black}
\end{spacing}
\label{tab:tradition}
\end{table*}

\begin{table*}[t]
\vspace{1em}
\centering
\caption{Ablation studies on the \dataset~test set. Constraining \textbf{the width of faces} and \textbf{the angle of ground truth gazes}, two comparisons are conducted. It is worthy of attention that, compared to the model that only predicts 3D gaze (\checkmark,$\times$,$\times$), the model that only predicts front-projection 2D gaze ($\times$,$F$,$\times$) achieves higher accuracy in the small-angle range (0-60\textdegree) and lower accuracy in the large-angle range due to the uneven distribution of the \textit{Gaze Sensitivity} (\cref{eq:gr}). This table shows the advantage of our full model with $F,T,S$ (\cref{eq:fproj}) and~$\mathcal{L}_{self}$~(\cref{eq:lself}).}
\begin{spacing}{1.15}
\arrayrulecolor[rgb]{0.902,0.902,0.902}
\begin{tabular}{|ccc||cccccccc|} 
\hline
\multirow{2}{*}{3D gaze task} & \multirow{2}{*}{2D gaze task} & \multirow{2}{*}{$\mathcal{L}_{self}$} & \multicolumn{8}{c|}{Gaze error (lower is better) \wrt \textbf{the width of faces}} \\ 
\cline{4-11}
 &  &  & 30-60 & 60-90 & 90-120 & 120-150 & 150-180 & 180-210 & 210-240 & \textgreater240 \\ 
\hline
\rowcolor[rgb]{0.902,0.902,0.902} \checkmark & $\times$ & $\times$ & 24.51 & 19.51 & 15.59 & 13.23 & 12.84 & 11.34 & 10.68 & 9.76 \\
$\times$ & $F$ & $\times$ & 24.00 & 18.73 & 14.56 & 12.6 & 12.22 & 10.83 & \underline{8.77} & 8.99 \\
\rowcolor[rgb]{0.902,0.902,0.902} \checkmark & $F$ & $F$ & \underline{23.44} & \underline{18.10} & \underline{14.16} & \underline{11.76} & \underline{11.09} & \underline{10.01} & 9.15 & \underline{8.09} \\
\checkmark & $F,T,S$ & $F,T,S$ & \textbf{22.94} & \textbf{17.55} & \textbf{13.69} & \textbf{11.08} & \textbf{11.13} & \textbf{9.41} & \textbf{8.17} & \textbf{7.74} \\ 
\hhline{|===::========|}
\multirow{2}{*}{3D gaze task} & \multirow{2}{*}{2D gaze task} & \multirow{2}{*}{$\mathcal{L}_{self}$} & \multicolumn{8}{c|}{Gaze error (lower is better) \wrt \textbf{the angle of GT gaze}} \\ 
\cline{4-11}
 &  &  & 0-20 & 20-30 & 30-40 & 40-50 & 50-60 & 60-70 & 70-80 & 80-90 \\ 
\hline
\rowcolor[rgb]{0.902,0.902,0.902} \checkmark & $\times$ & $\times$ & 9.12 & 9.94 & 10.83 & 11.59 & 14.3 & 18.67 & 26.86 & \textbf{43.04} \\
$\times$ & $F$ & $\times$ & 8.56 & \underline{8.49} & 9.75 & 10.77 & 12.64 & 18.37 & 30.2 & 54.35 \\
\rowcolor[rgb]{0.902,0.902,0.902} \checkmark & $F$ & $F$ & \underline{8.01} & 8.66 & \underline{9.21} & \underline{10.04} & \underline{11.52} & \underline{17.12} & \underline{26.57} & 43.75 \\
\checkmark & $F,T,S$ & $F,T,S$ & \textbf{7.97} & \textbf{8.45} & \textbf{8.99} & \textbf{9.51} & \textbf{10.53} & \textbf{16.36} & \textbf{23.31} & \underline{43.4} \\
\hline
\end{tabular}
\arrayrulecolor{black}
\end{spacing}
\label{tab:ablation}
\end{table*}

\subsection{Experimental Setup}

Our evaluation is mainly conducted on the test set of \dataset which is based on the validation set of Widerface \cite{yang2016wider} and ETH-XGaze \cite{zhang2020eth}, and we conduct gaze swap on 6277 faces of different resolutions. To match the input format of existing AGE methods, we conduct cropping and normalization on these faces. The experiments show that our method not only achieves higher accuracy and speed on synthetic data in \dataset, but we also test on real full images with gaze annotation from human experts and our method still performs better than existing AGE methods.

\subsection{Comparison with Existing \age~Methods\label{sec:exp_tradition}}

We compare our method with 4 full-face appearance based gaze estimation methods \cite{zhang2017s,zhang2020eth,cheng2021gaze}. They are all trained on the ETH-XGaze \cite{zhang2020eth} dataset to match the source of the gaze-swap of our test data which is also ETH-XGaze \cite{zhang2020eth}. As shown in \cref{tab:speed} and \cref{tab:tradition}, Full-face \cite{zhang2017s} is the earliest to be proposed and its speed is relatively high but it shows the worst accuracy. ETH-18 and ETH-50 \cite{zhang2020eth} are the models trained with ResNet18 and ResNet50 \cite{he2016deep} as backbones, where ETH-50 is used as the baseline method published in \cite{zhang2020eth}. GazeTR \cite{cheng2021gaze} is the latest method based on transformer design which achieves the highest accuracy among the four methods. However, it runs the slowest and cannot achieve real-time performance if there is more than one face in an image.

We test our method with MobileNet \cite{sandler2018mobilenetv2} and ResNet50 \cite{he2016deep} as backbones. The speed evaluation in \cref{tab:speed} shows that our method performs the best because our method can achieve similar speed as the RetinaFace \cite{deng2020retinaface}, the SOTA face detection method. Assuming that the four existing AGE methods mentioned above use RetinaFace \cite{deng2020retinaface} for face detection, their speed comparison with our method is clearly lagging behind even without considering the time spent on the normalization based on facial landmarks. In \cref{tab:speed}, the larger the \textit{\#face}, the slower the speed is. With the absolute speed advantage, \cref{tab:tradition} also shows that our method has the highest accuracy.

\subsection{Ablation study\label{sec:exp_ablation}}

\cref{tab:ablation} shows the ablation study of our method. Our full model contains 4 gaze tasks: one 3D gaze task (\textit{pitch}, \textit{yaw}) and three projected 2D gaze tasks (\textit{x}, \textit{y}) in which the projection in the front direction has a one-to-one correspondence with the 3D gaze, thus it can be easily transformed to \textit{pitch} and \textit{yaw} and compared with the ground truth. In addition, according to \cref{eq:lself}, we also add equal-loss to the 4 gaze predictions, which is also proven to be effective in \cref{tab:ablation}.

\begin{figure}[b]
\vspace{0.5em}
  \centering
  \includegraphics[width=\linewidth]{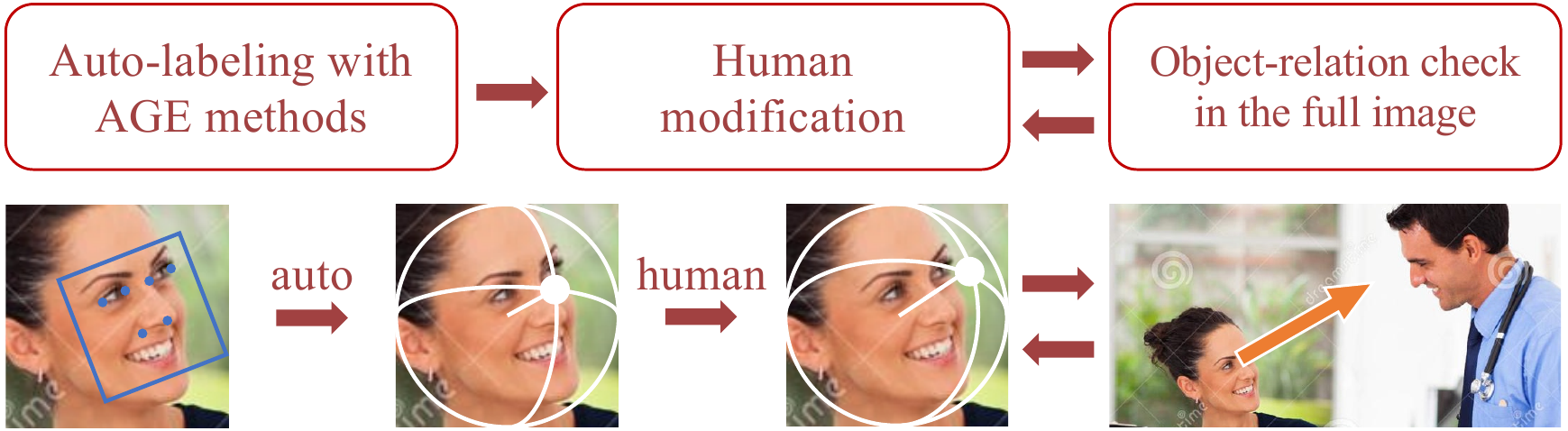}
  \vspace{0.5em}
  \caption{Human expert annotation pipeline. First, each face is cropped out, normalized, and then \cite{zhang2020eth} is employed to generate preliminary gaze labels. Second, human experts are asked to modify the gaze labels. Third, the gaze labels are shown in the original full image, and experts can modify them again according to object relations to get the final annotation.}
  \label{fig:annopipe}
\end{figure}

There are two parts in \cref{tab:ablation} constraining the target face size and ground truth gaze direction, respectively. For each table there are 4 rows of data. The first row indicates that the model only regresses 3D gaze; the second row indicates that the model only regresses 2D gaze projection in the front direction which is later converted to 3D gaze; the third row indicates that the model regresses both 3D gaze and 2D gaze projection (front) with a training loss restricting them to be equal; and the fourth row indicates that the model regresses all 4 gaze values and restricts them to be equal. The results in the table show the advantages of our full model. It is worthy of attention that, as we described in Method \cref{sec:proj}, compared to the model that only regresses 3D gaze, the model that only regresses 2D gaze in the front direction has higher accuracy in the small-angle range (0-60\textdegree) and lower accuracy in the large-angle range due to the uneven distribution of the \textit{Gaze Sensitivity} (\cref{eq:gr}). To solve this problem we propose to project 3D gaze from different directions, which solves the problem theoretically and achieves better results.

\begin{figure*}[t]
  \vspace{0.8em}
  \centering
  \includegraphics[width=\linewidth]{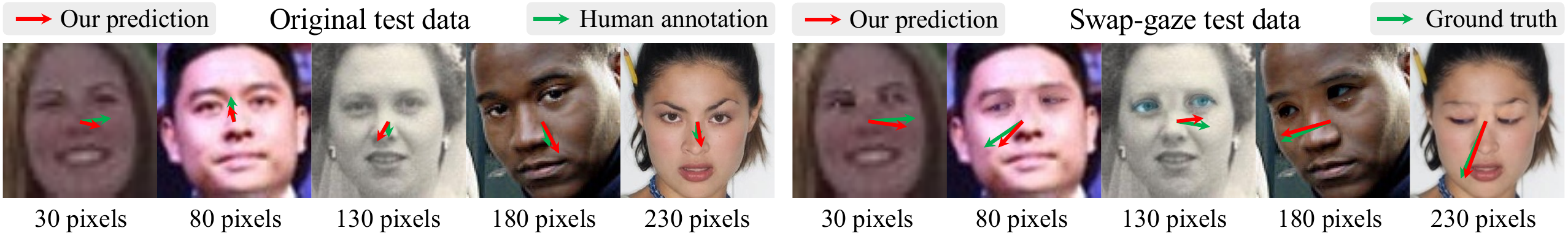}
  \caption{Visualization of predicted gaze on cropped single faces. Faces with various resolutions in human annotation dataset (left) and \dataset test set (right) are cropped from the whole images and resized for better visualization. Note that these results are generated by the MobileNet version of our model.}
  \label{fig:seesingle}
\end{figure*}

\begin{figure*}[t]
\vspace{0.8em}
  \centering
  \includegraphics[width=\linewidth]{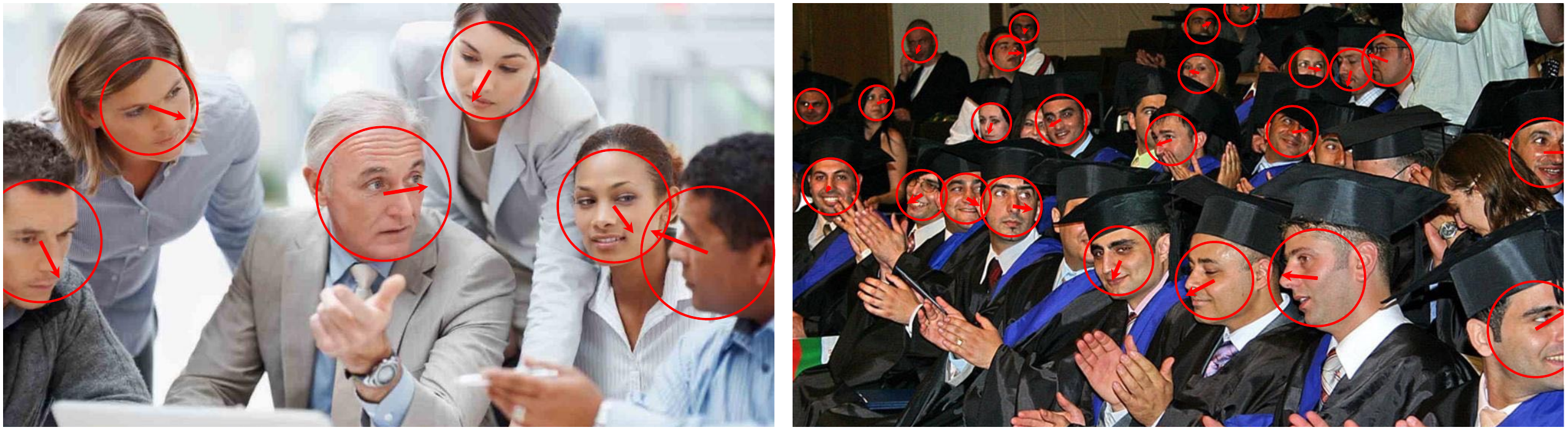}
  \caption{Visualization of multi-person gaze estimation on full images. The gaze directions of different people are estimated at the same time by the MobileNet version of our model.}
  \label{fig:seefull}
\end{figure*}

\subsection{Evaluation on Human Annotation Dataset\label{sec:exp_annotate}}

Besides testing on swap-gaze synthetic data, we also conduct evaluation on real full images. To get corresponding gaze labels, we ask some experts to conduct a subjective gaze annotation and finally acquire gaze annotation for 2719 faces in the Widerface \cite{yang2016wider} validation set. The annotation process is shown in \cref{fig:annopipe}. First, each face is cropped out for existing AGE methods to automatically generate preliminary gaze labels. Next, human experts are asked to modify the gaze labels and the gaze is then shown in the original full image so experts can modify them again according to object relations to get the final annotation. We also develop a GUI software to facilitate human experts to conduct the annotation.

\begin{table}[b]
\centering
\caption{Comparison on the human annotation dataset. Our methods (m: MobileNet0.25 backbone, r: ResNet50 backbone) show higher accuracy than existing AGE methods \cite{zhang2017s,zhang2020eth,cheng2021gaze} (trained on the ETH-XGaze \cite{zhang2020eth} dataset).}
\begin{spacing}{1.15}
\label{tab:annotest}
\arrayrulecolor[rgb]{0.902,0.902,0.902}
\begin{tabular}{|c||ccccc|} 
\hline
\multirow{2}{*}{Method} & \multicolumn{5}{c|}{Gaze error \wrt the width of faces} \\ 
\cline{2-6}
 & \small{0-60} & \small{60-120} & \small{120-180} & \small{180-240} & \small{\textgreater240} \\ 
\hhline{|=::=====|}
\rowcolor[rgb]{0.902,0.902,0.902} Full-face & 36.06 & 35.7 & 33.38 & 25.37 & 21.67 \\
ETH-18 & 31.95 & 31.38 & 28.5 & 22.85 & 19.93 \\
\rowcolor[rgb]{0.902,0.902,0.902} ETH-50 & 30.43 & 31.24 & 28.31 & \underline{22.11} & 18.79 \\
GazeTR & 36.00 & 33.53 & 31.10 & 26.81 & 23.59 \\
\rowcolor[rgb]{0.902,0.902,0.902} Ours-m & \underline{26.06} & \underline{25.88} & \underline{24.02} & 22.27 & \underline{18.41} \\
Ours-r & \textbf{25.90} & \textbf{22.69} & \textbf{22.02} & \textbf{19.92} & \textbf{15.54} \\
\hline
\end{tabular}
\arrayrulecolor{black}
\end{spacing}
\label{tab:annotest}
\end{table}

\cref{tab:annotest} shows the comparison between our method and existing AGE methods.
Although humans cannot perform well in 3D annotation tasks,
we can consider this experiment as a subjective test. The clear leading rank of our method shows our advantage. From another perspective, the relatively large test error shows the necessity of our proposed \dataset dataset with ground truth labels. We also show the visualization results in \cref{fig:seesingle} and \cref{fig:seefull}.

\section{Limitation and Future Work}

\vspace{-0.5em}
First, our method cannot produce estimates for people who show their backs to the camera or look towards the back side of the scene. This is a common problem for existing appearance-based gaze estimation methods while it is inevitable in the real world. In future research, such back-to-camera situations can be further considered and tried to handle.
Second, although we propose an effective method to synthesize full images with multi-person gaze ground truth, it could be still worth considering to try to collect real data directly with accurate gaze directions of multiple people in the wild.

\section{Conclusion}
\vspace{-0.5em}
We propose the first one-stage gaze estimation method, \ie, \method, which can estimate multi-user gaze directions simultaneously in a full image. In addition, we design a projection-based self-supervised strategy that can further improve the gaze accuracy.
To enable one-stage gaze estimation training and evaluation, we provide a new gaze dataset, \dataset, which is generated by a sophisticated swap-gaze procedure to produce full images of multi-person with gaze ground truth data. Finally, our method outperforms state-of-the-art methods in terms of gaze accuracy and speed.

{\small
\bibliographystyle{ieee_fullname}
\bibliography{egbib}
}

\end{document}